\documentclass[10pt,twocolumn,letterpaper]{article}

\usepackage{cvpr}              
\usepackage[accsupp]{axessibility}
\usepackage{graphicx}
\usepackage{amsmath}
\usepackage{amssymb}
\usepackage{booktabs}
\usepackage{multirow}
\usepackage{makecell}
\usepackage{adjustbox}
\usepackage{cuted}
\def\eg{\emph{e.g}\onedot} 
\def\ie{\emph{i.e}\onedot} 
 
\usepackage{xcolor}

\usepackage{pifont}
\definecolor{citecolor}{HTML}{0071BC}
\usepackage[pagebackref,breaklinks,colorlinks,citecolor=citecolor]{hyperref}

\usepackage[capitalize]{cleveref}
\crefname{section}{Sec.}{Secs.}
\Crefname{section}{Section}{Sections}
\Crefname{table}{Table}{Tables}
\crefname{table}{Tab.}{Tabs.}

\def\cvprPaperID{9200} 
\def\confName{CVPR}
\def\confYear{2023}

\begin{document}

\title{TriVol: Point Cloud Rendering via Triple Volumes}

\author{
 Tao Hu $^{1}$$^*$ \quad Xiaogang Xu$^{1}$$^*$ \quad  Ruihang Chu $^1$ \quad Jiaya Jia$^{1,2}$\\
 $^1$ The Chinese University of Hong Kong \qquad
 $^2$ SmartMore\\
 {\tt\small \{taohu,xgxu,rhchu,leojia\}@cse.cuhk.edu.hk}
 \vspace{-0.4in}
}

\maketitle

\begin{strip}
\centering
\includegraphics[width=0.85\textwidth]
{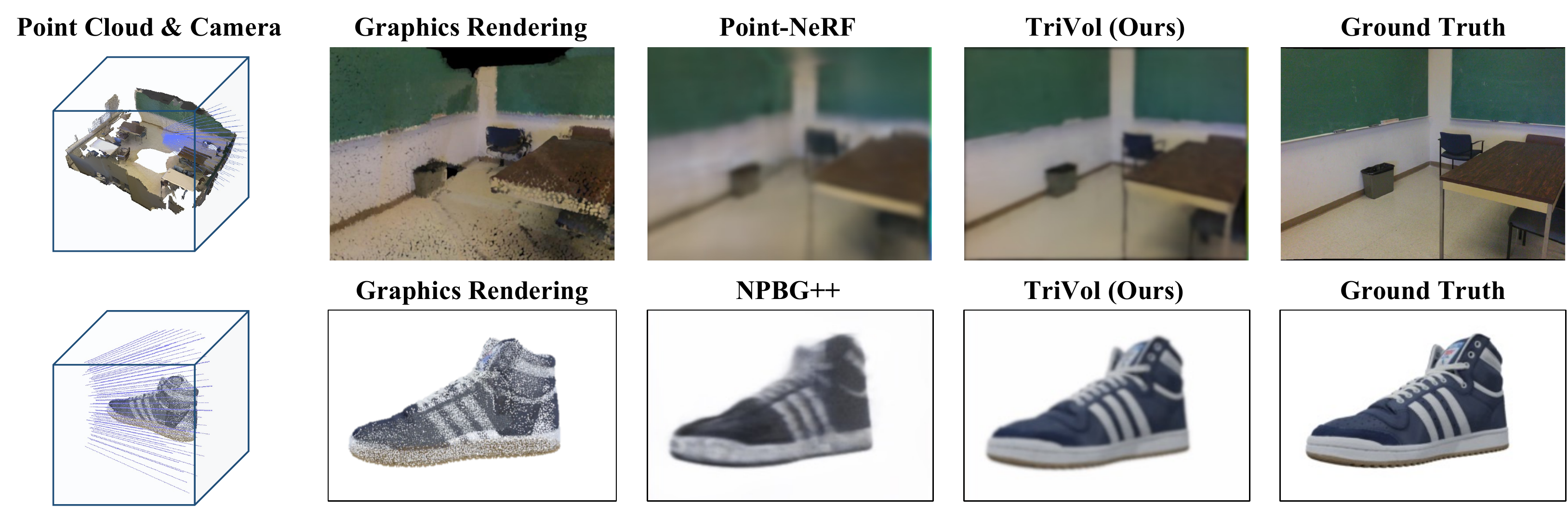}
\vspace{-0.1in}
\captionof{figure}{
Given the colored point cloud of a category-specific scene or object, our TriVol can render photo-realistic images.
}
\label{fig:teaser}
\end{strip}


\renewcommand{\thefootnote}{\fnsymbol{footnote}}
\footnotetext[1]{Equal Contribution.}

\begin{abstract}
Existing learning-based methods for point cloud rendering adopt various 3D representations and feature querying mechanisms to alleviate the sparsity problem of point clouds. However, artifacts still appear in rendered images, due to the challenges in extracting continuous and discriminative 3D features from point clouds. In this paper, we present a dense while lightweight 3D representation, named TriVol, that can be combined with NeRF to render photo-realistic images from point clouds. Our TriVol consists of triple slim volumes, each of which is encoded from the point cloud. TriVol has two advantages. First, it fuses respective fields at different scales and thus extracts local and non-local features for discriminative representation. Second, since the volume size is greatly reduced, our 3D decoder can be efficiently inferred, allowing us to increase the resolution of the 3D space to render more point details. Extensive experiments on different benchmarks with varying kinds of scenes/objects demonstrate our framework's effectiveness compared with current approaches. Moreover, our framework has excellent generalization ability to render a category of scenes/objects without fine-tuning. The source code is available at \url{https://github.com/dvlab-research/TriVol.git}.


\end{abstract}

\section{Introduction}
\label{sec:introduction}

Photo-realistic point cloud rendering (without hole artifacts and with clear details) approaches can be employed for a variety of real-world applications, \eg, the visualization of automatic drive~\cite{mallya2020world,chen2021geosim,cortinhal2021semantics,NPLF}, digital humans~\cite{nguyen2022free,NPBG,kopanas2021point}, and simulated navigation~\cite{wijmans2019embodied,devaux2016realtime,chaurasia2013depth}. Traditional point cloud renderers~\cite{Pytorch3d} adopt graphics-based methods, which do not require any learning-based models. They project existing points as image pixels by rasterization and composition. However, due to the complex surface materials in real-world scenes and the limited precision of the 3D scanners, there are a large number of missing points in the input point cloud, leading to vacant and blurred image areas inevitably as illustrated in Fig.~\ref{fig:teaser}.

In recent years, learning-based approaches~\cite{NPBG,NPBG++,PointNeRF,point_based_radiance_fields,NPCR} have been proposed to alleviate the rendering problem in graphics-based methods. 
They use a variety of querying strategies in the point cloud to obtain continuous 3D features for rendering, \eg, the ball querying employed in the PointNet++~\cite{PointNet++} and the KNN-based querying in Point-NeRF~\cite{PointNeRF}. 
However, if the queried position is far away from its nearest points, the feature extraction process will usually fail and have generalization concerns.
To guarantee accurate rendering, two groups of frameworks are further proposed. 
The first group~\cite{ADOP,NPBG,NPBG++} projects the features of all points into the 2D plane and then trains 2D neural networks, like UNet~\cite{UNet}, to restore and sharpen the images. However, since such a 2D operation is individual for different views, the rendering outcomes among nearby views are inconsistent~\cite{Epigraf,StyleNeRF}, \ie, the appearance of the same object might be distinct under different views.
To overcome the artifact, physical-based volume rendering \cite{volume_rendering} in Neural Radiance Fields (NeRF) \cite{NeRF} without 2D neural rendering is an elegant solution \cite{Epigraf}. 
The second group~\cite{NPCR,Structural_Textural_Representation} applies a 3D Convolutional Network (ConvNet) to extract a dense 3D feature volume, then acquires the 3D feature of arbitrary point by trilinear interpolation for subsequent volume rendering.
Nevertheless, conducting such a dense 3D network is too heavy for high-resolution 3D representation, limiting their practical application. In summary, to the best of our knowledge, there is currently no lightweight point cloud renderer whose results are simultaneously view-consistent and photo-realistic.

In this paper, we propose a novel 3D representation, named TriVol, to solve the above issues. Our TriVol is composed of three slender volumes which can be efficiently encoded from the point cloud. Compared with dense grid voxels, the computation on TriVol is efficient and the respective fields are enlarged, allowing the 3D representation with high resolution and multi-scale features. Therefore, TriVol can provide discriminative and continuous features for all points in the 3D space.
By combining TriVol with NeRF \cite{NeRF}, the point cloud rendering results show a significant quantitative and qualitative improvement compared with the state-of-the-art (SOTA) methods.

In our framework, we develop an efficient encoder to transform the point cloud into the \textit{Initial TriVol} and then adopt a decoder to extract the \textit{Feature TriVol}. 
Although the encoder can be implemented with the conventional point-based backbones~\cite{PointNet,PointNet++}, we design a simple but effective grouping strategy that does not need extra neural models. The principle is first voxelizing the point cloud into grid voxels and then re-organizing the voxels on each of three axes individually, which is empirically proven to be a better method. 
As for the decoder, it can extract the feature representation for arbitrary 3D points. Hence, we utilize three independent 3D ConvNet to transfer each volume into dense feature representations. Due to the slender shape of the volumes, the computation of the 3D ConvNet is reduced. Also, the 3D ConvNet can capture more non-local information in the grouped axis via a larger receptive field, and extract local features in the other two directions.

With the acquired dense \textit{Feature TriVol}, the feature of any 3D points can be queried via trilinear interpolation. By combining the queried features with the standard NeRF \cite{NeRF} pipeline, the photo-realistic results are achieved. Extensive experiments are conducted on three representative public datasets (including both datasets of scene~\cite{ScanNet} and object ~\cite{ShapeNet,GSO}) level, proving our framework's superiority over recent methods.
In addition, benefiting from the discriminative and continuous feature volume in TriVol, our framework has a remarkable generalization ability to render unseen scenes or objects of the same category, when there is no further fine-tuning process.

In conclusion, our contributions are three-fold.
\begin{itemize}
    \item 
    We propose a dense yet efficient 3D representation called TriVol for point cloud rendering. It is formed by three slim feature volumes and can be efficiently transformed from the point cloud.
    \item We propose an effective encoder-decoder framework for TriVol representation to achieve photo-realistic and view-consistent rendering from the point cloud.
    \item Extensive experiments are conducted on various benchmarks, showing the advantages of our framework over current works in terms of rendering quality.
\end{itemize}

\section{Related Works}
\label{sec:related_works}

\subsection{3D Representation}
3D representation is very important when analyzing and understanding 3D objects or scenes. There are several important 3D representations, including point cloud, dense voxels~\cite{NPCR}, sparse voxels~\cite{ME}, Multi-Plane Images (MPI), triple-plane (triplane)~\cite{Epigraf,eg3d,noguchi2022unsupervised,gao2022get3d}, multi-plane~\cite{srinivasan2019pushing,mildenhall2019local,shih20203d}, and NeRF~\cite{NeRF}, designed from different tasks. A point cloud is a set of discrete data points in space representing a 3D object or scene. Each point location has a coordinate value and could further contain the color. The point cloud is an efficient 3D representation that is usually captured from a real-world scanner or obtained via Multi-view Stereo (MVS) ~\cite{MVS}.
Each voxel in the dense voxels represents a value on a regular grid in the 3D space. By using interpolation, the continuous 3D features of all 3D positions can be obtained.
The sparse voxels \cite{Sparse_3D_CNN,SparseConvNet,ME} are the compressive representation of the dense voxels since only parts of the voxels have valid values. 
The triplane representation~\cite{Epigraf,eg3d} is also the simplification of dense voxels, obtained by projecting the 3D voxels to three orthogonal planes. 
The MPI~\cite{srinivasan2019pushing,mildenhall2019local,NPCR} represents the target scene as a set of RGB and alpha planes within a reference view frustum.
Moreover, NeRF~\cite{NeRF} is a recently proposed implicit 3D representation, which can represent the feature of any input 3D coordinate with an MLP. The MLP maps the continuous input 3D coordinate to the geometry and appearance of the scene at that location. We propose TriVol as a new 3D representation of the point cloud and demonstrate its advantages in point cloud rendering.

\subsection{Point-based Rendering}
Point cloud rendering can be implemented with graphics- and learning-based approaches.
The points are projected to the 2D plane via rasterization and composition in the graphics-based algorithms~\cite{Pytorch3d}. The learning-based methods ~\cite{kopanas2021point,lassner2021pulsar,ME,PointNeRF} design various strategies to compensate the missing information from the sparse point cloud.
For example, ME~\cite{ME} first conducts sparse ConvNet to extract features for existing points, then computes the features of arbitrary 3D points by ball querying in the local space. Obviously, most of the points in the whole 3D space have no meaningful features.
Point-NeRF~\cite{PointNeRF} makes use of multi-view images to enhance the features of the input point cloud, formulating the sparse 3D feature volume and then querying any point features by $K$ Nearest Neighbors (KNN)~\cite{PointNet++}.
Also, quite a few learning-based methods ~\cite{wiles2020synsin,pumarola2020c,prokudin2021smplpix,zakharkin2021point,ADOP,NPBG,NPBG++} project the point cloud onto the 2D plane and utilize the 2D networks to recover the hole artifacts caused by the point cloud's discrete property.
For instance, NPBG~\cite{NPBG} renders the point
cloud with learned neural features in multiple scales and sets a 2D UNet for refinement.
Furthermore, several approaches construct 3D feature volumes for rendering~\cite{NPCR}, \eg, NPCR~\cite{NPCR} uses 3D ConvNet to obtain 3D volumes from point clouds and produce multiple depth layers to synthesize novel views. 

\begin{figure*}[t]
	\centering
	\includegraphics[width=0.85\linewidth]{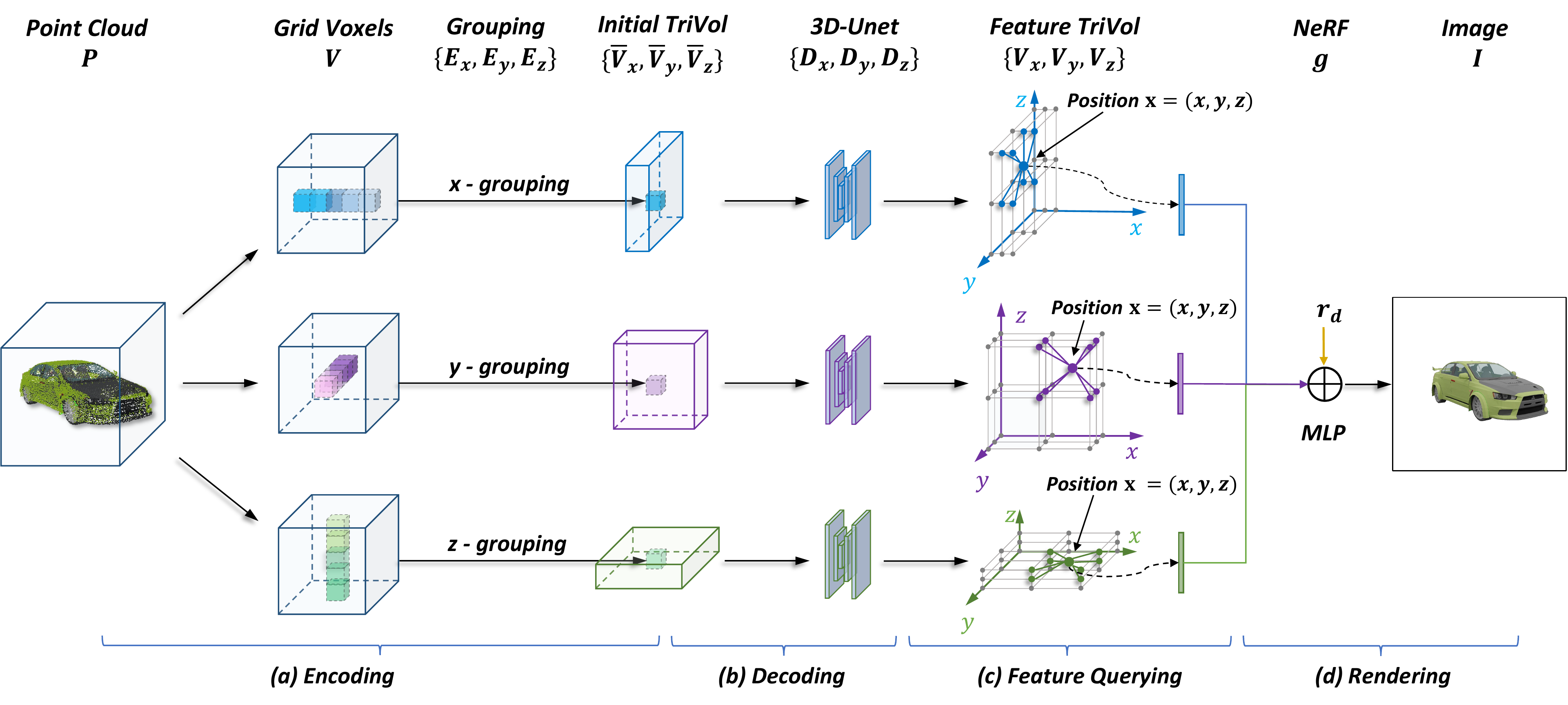}
	\vspace{-0.1in}
	\caption{Overview of the proposed TriVol for point cloud rendering. \textbf{(a) Encoding}: Input point cloud is encoded to our \textit{Initial TriVol} along $x$, $y$, and $z$ axis; \textbf{(b) Decoding}: Each volume is decoded to dense feature volume via a unique 3D UNet; \textbf{(c) Feature Querying}: Any point's feature is queried by the trilinear interpolation in the \textit{Feature TriVol}; \textbf{(d) Rendering}: We combine the queried point feature with NeRF to render the final image.}
    \label{fig:overview}
    \vspace{-0.1in}
\end{figure*}

\section{Approach}
\label{sec:approach}
We aim to train a category-specific point renderer $\mathcal{R}$ to directly generate photo-realistic images $I$ (the image height and width are denoted as $H$ and $W$) from the colored point cloud $P$, given camera parameters (intrinsic parameter $K$ and extrinsic parameters $R$ and $t$). When rendering novel point clouds of the same category, no fine-tuning process is required. 
The rendering process can be represented as 
\begin{equation}
    I = \mathcal{R}(P | R, t, K),
    \label{eq:rendering}
\end{equation}
where $P$ is usually obtained from MVS~\cite{MVS,openmvs2020}, LiDAR scanners~\cite{ScanNet}, or sampled from synthesized mesh models. In this section, we first encode the point cloud as the proposed TriVol, then utilize three 3D UNet to decode it into the feature representation. Finally, we combine NeRF \cite{NeRF} by querying point features from TriVol at sampled locations to render the final images. An overview of our method is illustrated in Fig. \ref{fig:overview}.

\subsection{TriVol Representation}
\label{sec:representation}

\noindent\textbf{Grid Voxels.} To begin with, we voxelize sparse point cloud $P$ into grid voxels $V$ with shape $\mathbb{R}^{C\times S\times S\times S}$, where $S$ is the resolution of each axis, and $C$ is the number of feature channels.
Since $V$ is a sparse tensor, directly querying within $V$ will only get meaningless values for most of the 3D locations, leading to the vacant areas in the rendered images.
Therefore, the critical step is transforming the sparse $V$ into a dense and discriminative representation.
One approach is employing a 3D encoder-decoder, \eg, 3D UNet.
Nevertheless, such a scheme is not efficient to represent a high-resolution space and render fine-grained details. The reason is two-fold: 1) conducting 3D ConvNet on $V$ requires a lot of computations and memory resources, leading to a small value of $S$ \cite{NPCR}; 2) the sparsity of $V$ impedes the feature propagation since regular 3D ConvNet only has a small kernel size and receptive field.

\noindent\textbf{From grid voxels to TriVol.} 
To overcome above two issues of $V$, we propose TriVol, including ${V_x, V_y, V_z }$, as a novel 3D representation. As illustrated in Fig. \ref{fig:overview}, each item in TriVol is a thin volume whose resolution of one axis is obviously smaller than $S$, and the others are the same as $S$. As a consequence, the number of total voxels is reduced for lightweight computations.
Note that our TriVol is different from triple-plane representations that are employed in existing works, \eg, ConvOnet~\cite{convonet} and EpiGRAF \cite{Epigraf}. Their point features are projected to three standard planes, thus much 3D information might be lost \cite{convonet}.

\subsubsection{Encoder for Initial TriVol}
\label{sec:encoder}
We first encode the input point cloud into the \textit{Initial TriVol} ($\{\bar{V}_x, \bar{V}_y, \bar{V}_z \}$). This process can be completed by existing point-cloud-based networks, such as PointNet\cite{PointNet}, PointNet++ \cite{PointNet++}, and Dense/Sparse ConvNet \cite{ME,Sparse_3D_CNN}.
Nevertheless, these networks bring an additional and heavy computation burden. Instead, we design an efficient strategy without an explicit learning model. 

The main step in our encoder is the x-grouping, y-grouping, and z-grouping along different axes. 
The procedure can be denoted as $\{\bar{V}_x, \bar{V}_y, \bar{V}_z \}=E(V)$, where $E=\{E_x, E_y, E_z \}$. Specifically, to obtain the slim volumes of the $x$-axis, we first divide $V$ into $G \times S \times S$ groups along the $x$-axis, thus each group contains $N=S/G$ voxels. Then we concatenate all $N$ voxels in each group as one new feature voxel to obtain $\bar{V}_{x}\in\mathbb{R}^{(C\cdot N)\times G\times S \times S}$, where $C\cdot N$ is the number of feature channels for each new voxel. $\bar{V}_y$ and $\bar{V}_z$ are encoded by the similar grouping method but along $y$ and $z$ axis, respectively.
Therefore, $E$ can be formulated as
\begin{equation}
\small
\begin{aligned}
\bar{V}_{x}=E_x(V) \in\mathbb{R}^{(C\cdot N)\times G\times S \times S}\\
\bar{V}_{y}=E_y(V)\in\mathbb{R}^{(C\cdot N)\times S\times G \times S}\\
\bar{V}_{z}=E_z(V)\in\mathbb{R}^{(C\cdot N)\times S\times S \times G}\\
\end{aligned}
    \label{eq:grouping}
\end{equation}

Our encoder is simple and introduces two benefits. Firstly, we can set the different sizes of $G$ and $S$ to balance the performance and computation. When $G \ll S$, huge computing resources are not required, compared with grid voxels $V$. Thus, we can increase the resolution $S$ to model more point cloud details. Secondly, since the voxels in the same group share the identical receptive field, the receptive field on the grouped axis is amplified $N$ times, allowing the subsequent decoder to capture global features on the corresponding axis without using a large kernel size.
For instance, in volume $\bar{V}_x$, we can extract non-local features on the $x$-axis and local features on the $y$ and $z$ axis.

\subsubsection{Decoder for Feature TriVol}
\label{sec:decoder}
After obtaining the \textit{Initial TriVol} $\{\bar{V}_x, \bar{V}_y, \bar{V}_z \}$ from the encoder, we utilize 3D ConvNet to decode them as the \textit{Feature TriVol} $\{V_x, V_y, V_z \}$.
Our \textit{Feature TriVol} decoder consists of three 3D UNet \cite{UNet} modules $D=\{D_x, D_y, D_z \}$. Each 3D UNet can acquire non-local features on the grouped axis with the amplifying receptive field and can extract local features on the ungrouped two axes that preserve the standard local receptive field.
The decoding procedure can be represented as
\begin{equation}
    V_x=D_{x}(\bar{V}_x), \; V_y=D_y(\bar{V}_y), \; V_z=D_z(\bar{V}_z),
    \label{eq:decoder}
\end{equation}
where $V_x \in \mathbb{R}^{F\times G \times S \times S}$, $V_y \in \mathbb{R}^{F\times S \times G \times S}$, and $V_z \in \mathbb{R}^{F\times S \times S \times G}$, and $F$ denotes channel number of \textit{Feature TriVol}.
Although three 3D UNets are required, the small number of $G$ still makes it possible for us to set a large resolution $S$ without increasing computing resources (verified in Sec.~\ref{sec:experiments}), resulting in realistic images with rich details.

\subsection{TriVol Rendering}
\label{sec:rendering}
The encoder and decoder modules have transformed the sparse point cloud into a dense and continuous \textit{Feature TriVol}. Therefore, the feature of any 3D location can be directly queried by the trilinear interpolation in $\{V_x,V_y,V_z\}$.
Finally, the rendering images can be obtained from the point cloud by following feature querying and volume rendering pipeline of NeRF~\cite{NeRF}.

\subsubsection{Feature Querying}
The feature querying consists of point sampling along the casting ray and feature interpolation in the TriVol.

\vspace{2mm}
\noindent\textbf{Point sampling}. 
Given the camera parameters $\{R, t, K\}$, we can calculate a random ray with camera center $\textbf{r}_o \in \mathbb{R}^3$ and normalized direction $\textbf{r}_d \in \mathbb{R}^3$, we adopt the same coarse-to-fine sampling strategy as NeRF \cite{NeRF} to collect the queried points $\textbf{x} \in \mathbb{R}^3$ along the ray, as
\begin{equation}
    \textbf{x} = \textbf{r}_o + z \cdot \textbf{r}_d, \quad z \in [z_n, z_f],
    \label{eq:sampling}
\end{equation}
where $z_n$, $z_f$ are the near and far depths of the ray.

\vspace{2mm}
\noindent\textbf{Querying}. For \textit{Feature TriVol} $\{V_x, V_y, V_z \}$ and a queried location $\textbf{x}$, 
we first utilize trilinear interpolation to calculate 3 feature vectors: $V_x(\textbf{x})\in \mathbb{R}^{F}$, $V_y(\textbf{x}) \in \mathbb{R}^{F}$, and $V_z(\textbf{x}) \in \mathbb{R}^{F}$ as shown in Fig.~\ref{fig:overview}. Then, we concatenate them as the final feature $F(\textbf{x})$ for location $\textbf{x}$, as
\begin{equation}
    F(\textbf{x})= V_x(\textbf{x}) \oplus V_y(\textbf{x}) \oplus V_z(\textbf{x}),
\end{equation}
where $\oplus$ is the concatenation operation.

\subsubsection{Volume Rendering}
\noindent\textbf{Implicit mapping}.
For the queried feature of all points on the ray, we set a Multi-Layer Perceptron (MLP) as an implicit function $g$ to map interpolated feature $F(\textbf{x})$ to their densities $\sigma \in \mathbb{R}_+$ and colors $c \in \mathbb{R}^3$, with the view direction $\textbf{r}_d$ as the condition, as
\begin{equation}
    \sigma, \textbf{c} = g(F(\textbf{x}), \textbf{r}_d).
\end{equation}


\begin{table*}[t]
\begin{adjustbox}{width=0.90\linewidth,center}
\begin{tabular}{c|cccc}
\hline
\textbf{Methods} & \textbf{3D Representation} & \textbf{Feature Extraction} & \textbf{Feature Query} & \textbf{Rendering} \\ \hline
\textbf{NPBG++ \cite{NPBG++}} & Point Cloud & 2D UNet & - & Graphics \& CNN \\ 
\textbf{Dense 3D ConvNet} & Grid Voxels & 3D UNet & Trilinear Interpolation & NeRF \\ 
\textbf{Sparse 3D ConvNet} & Sparse Voxels & 3D Sparse UNet & Ball Query & NeRF \\ 
\textbf{Point-NeRF \cite{Point-NeRF}} & Point Cloud & MLP & KNN & NeRF\\
\textbf{Ours} & Triple Volumes (TriVol) & 3D UNet & Trilinear Interpolation & NeRF \\ \hline
\end{tabular}
\end{adjustbox}
\vspace{-0.1in}
\caption{Compare different point cloud renderers in 3D representation, feature extraction, feature query, and rendering strategy.}
\label{tab:method_details}
\vspace{-0.1in}
\end{table*}

\noindent\textbf{Rendering}.
The final color of each pixel $\hat{\textbf{c}}$ can be computed by accumulating the radiance on the ray through the pixel using volume density \cite{volume_rendering}.
Suppose there are $M$ points on the ray, such a volume rendering can be described as
\begin{align}
	\begin{aligned}
	    &\hat{\textbf{c}} = \sum_{i=1}^{M} T_i \alpha_i \textbf{c}_i, \\ 
	&\alpha_i = 1 -\textbf{ exp}(-\sigma_i \delta_i), \\ 
	&T_i = \textbf{exp}(-\sum_{j=1}^{i-1}\sigma_j\delta_j),
	\end{aligned}
	\label{eq:volume_rendering}
\end{align}
where $T_i$ represents volume transmittance, $\delta_j$ is the distance between neighbor samples along the ray $\textbf{r}$.

\subsection{Loss Function}
Our model is only supervised by the rendering loss, which is calculated by the mean square error between rendered colors and ground-truth colors, as
\begin{equation}
    \begin{aligned}
    &\{\hat{\textbf{c}}_1, ..., \hat{\textbf{c}}_{H\times W}\}=\mathcal{R}(P| R, t, K),\\
    &\mathcal{R} = \{D_x, D_y, D_z, g \},\\
    &\mathcal{L} = \lVert  \hat{\textbf{c}}_i - \bar{\textbf{c}}_i  \rVert_2^2,\\
    \end{aligned}
\end{equation}
where $I=\{\hat{\textbf{c}}_1, ..., \hat{\textbf{c}}_{H\times W}\}$ are the rendered colors, $\{\bar{\textbf{c}}_1, ..., \bar{\textbf{c}}_{H\times W}\}$ are the ground-truth colors.

\begin{table*}[h]
\begin{adjustbox}{width=1.0\linewidth,center}
\begin{tabular}{l|ccc|ccc|ccc}
\hline
  & \multicolumn{3}{c|}{ScanNet \cite{ScanNet}} & \multicolumn{3}{c|}{ShapeNet \cite{ShapeNet}} & \multicolumn{3}{c}{Google Scanned Objects \cite{GSO}} \\
 & PSNR ($\uparrow$) & SSIM ($\uparrow$) & LPIPS ($\downarrow$) & PSNR ($\uparrow$) & SSIM ($\uparrow$) & LPIPS ($\downarrow$) & PSNR ($\uparrow$) & SSIM ($\uparrow$) & LPIPS ($\downarrow$) \\ \hline
Graphics Renderer \cite{Pytorch3d} & 13.62 & 0.528 & 0.779 & 19.24 & 0.814 & 0.182 & 23.14 & 0.829 & 0.153 \\
Sparse-ConvNet \cite{ME} & 15.27 & 0.646 & 0.602 & 22.16 & 0.836 & 0.159 & 25.36 & 0.868 & 0.128 \\
NPCR \cite{NPCR} & 16.22 & 0.659 & 0.574 & 23.41 & 0.855 & 0.136 & 27.73 & 0.892 & 0.107 \\
ConvONet \cite{convonet} & 16.43 & 0.665 & 0.584 & 24.25 & 0.848 & 0.122 & 28.17 & 0.917 & 0.093 \\
ADOP \cite{ADOP} & 16.83  & 0.699 & 0.577  & 24.96  & 0.857  & 0.129 & 29.06 & 0.922 & 0.089 \\
NPBG++ \cite{NPBG++} & 16.81 & 0.671 & 0.585 & 25.32 & 0.874 & 0.120 & 29.42 & 0.929 & 0.081 \\
Point-NeRF \cite{Point-NeRF} & 17.53 & 0.685 & 0.517 & \textbf{25.73} & \textbf{0.897} & \textbf{0.107} & \textbf{29.84} & \textbf{0.938} & \textbf{0.069} \\ 
Voxels-128  & \textbf{17.65} & \textbf{0.694} & \textbf{0.538} & 25.53 & 0.872 & 0.116 & 29.41 & 0.926 & 0.078  \\ \hline
Ours & \textbf{18.56} & \textbf{0.734} & \textbf{0.473} & \textbf{27.22} & \textbf{0.927} & \textbf{0.084} & \textbf{31.24} & \textbf{0.961} & \textbf{0.045} \\ \hline
\end{tabular}
\end{adjustbox}
\vspace{-0.1in}
\caption{Quantitative comparison for point cloud rendering accuracy between ours and the state-of-the-art methods as well as baselines on the ScanNet \cite{ScanNet}, ShapeNet \cite{ShapeNet}, and Google Scanned Objects \cite{GSO} datasets.}
\label{tab:metric_comparison}
\vspace{-0.1in}
\end{table*}

\section{Experiments}
\label{sec:experiments}
\subsection{Datasets}
We evaluate the effectiveness of our framework with the TriVol representation on object-level and scene-level datasets. 
For the object level, we use both synthesized and real-world scanned datasets, including \textbf{ShapeNet}~\cite{ShapeNet} and \textbf{Google Scanned Objects} (GSO)~\cite{GSO}.
ShapeNet is a richly-annotated and large-scale 3D synthesized dataset. There are about 51,300 unique 3D textured mesh models, and we choose the common \textit{Car} category for evaluation. 
GSO dataset contains over 1000 high-quality 3D-scanned household items, and we perform experiments on the category of \textit{shoe}. 
The point clouds in object-level datasets can be obtained by 3D mesh sampling~\cite{mesh-sampling}, and ground-truth rendered images are generated by Blender~\cite{blender} under random camera poses. 
For the scene level, we conduct the evaluation on the \textbf{ScanNet}~\cite{ScanNet} dataset, which contains over 1500 indoor scenes. Each scene is constructed from an RGBD camera. We split the first 1,200 scenes as a training set and the rest as
a testing set. 

\subsection{Implementation Details}
\label{sec:details}
We set the volume resolution $S$ to 256, and the number of groups $G$ to 16. The decoders ($D_x$, $D_y$, and $D_z$) do not share the weights. The number of layers in NeRF's MLP $g$ is 4. For each iteration, we randomly sample $1024$ rays from one viewpoint. The number of points for each ray is $64$ for both coarse and fine sampling. The resolutions $H\times W$ of rendered images on the ScanNet \cite{ScanNet} dataset are $512\times 640$, and the other datasets are $256\times 256$. AdamW \cite{AdamW} is adopted as the optimizer, where the learning rate is initialized as $0.001$ and will decay to $0.0001$ after 100 epochs. We train all the models using four RTX 3090 GPUs. For more details, please refer to the supplementary file.

\subsection{Baselines}
Besides the comparison with existing point cloud rendering methods, \eg, NPBG++~\cite{NPBG++}, ADOP \cite{ADOP} and Point-NeRF~\cite{Point-NeRF}, several important baselines are also combined with NeRF \cite{NeRF} to demonstrate the effectiveness of our framework. They can be described as follows:
\begin{itemize}

\item \textbf{Voxels-128}: This baseline generates dense voxels with 3D dense UNet. However, due to its constrained efficiency and the memory limitation of the computation resource, the voxel resolution is set as $128^3$.
\vspace{-0.05in}

\item \textbf{Sparse ConvNet}: MinkowskiEngine~\cite{ME} is a popular sparse convolution library. We make use of its sparse 3D UNet, \ie, MinkUnet34C~\cite{ME}, as the baseline.
Its feature querying is performed by a ball query.
\vspace{-0.05in}
\item \textbf{ConvOnet \cite{convonet}}: ConvOnet converts the point cloud features into a triple-plane representation for 3D reconstruction. We replace its occupancy prediction with NeRF module to achieve point cloud rendering.
\vspace{-0.05in}
\end{itemize}
Note that Voxels-128 has the same channel number as TriVol. One alternative baseline is the dense voxels with high resolution, e.g., 256, and reduced channel number for efficiency. However, the minor channel number will impede the formulation of discriminative feature volumes, and such an alternative baseline has been proved to have terrible rendering results empirically.
The differences between our framework and these baselines are summarized in Tab.~\ref{tab:method_details}. We adopt the metrics of PSNR, SSIM~\cite{wang2004image}, and LPIPS~\cite{LPIPS} for evaluation.

\subsection{Evaluate Scene-Level Rendering}
\begin{figure*}[t]
	\centering
	\includegraphics[width=0.8\linewidth]{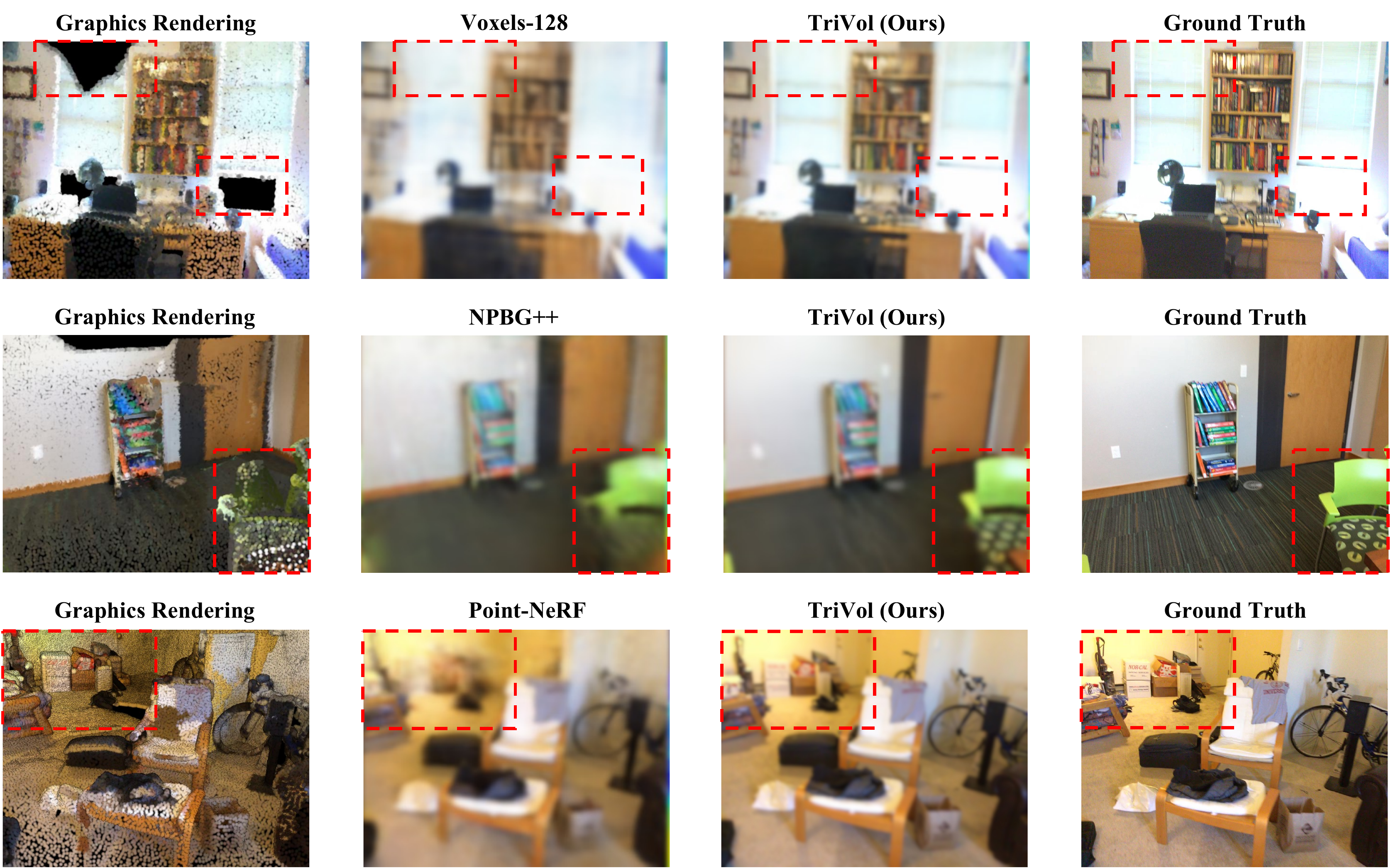}
	\caption{Qualitative point cloud rendering comparison between ours and SOTA methods and baselines on the ScanNet dataset.}
    \label{fig:visualization-scannet}
\end{figure*}

In this experiment, we compare ours with baselines, and SOTA methods \cite{NPCR,Point-NeRF,NPBG++} on the ScanNet dataset \cite{ScanNet}. For each scene, we sample 100k points on the provided textured mesh as the colored point cloud. 
Quantitative and qualitative results are presented in Tab. \ref{tab:metric_comparison} and Fig. \ref{fig:visualization-scannet}, respectively.

The point clouds at the scene level usually have missing points and parts, which causes hole artifacts in graphics-based rendering images, as shown in the left column of Fig. \ref{fig:visualization-scannet}. The traditional voxel-based method could not generate high-quality images due to its low-resolution representation. There are many artifacts in NPBG++ \cite{NPBG++}, since the 2D-CNN-based rendering only employs the limited 2D information, not 3D context, to remove the discrete issues, leading to imprecise and unrealistic results when there is no fine-tuning stage.
Moreover, the view-inconsistent shortcomings of 2D-CNN-based approaches will be demonstrated in the supplementary file.
Furthermore, Point-NeRF \cite{Point-NeRF} could not recover the missing parts. The reason is that the KNN strategy will usually fail if the nearest neighbor points are far from the queried points \cite{PointNet++}. Ours performs the best in complementing the missing areas and enhancing local details, thus could render photo-realistic images. 
Moreover, as shown in Tab.~\ref{tab:metric_comparison}, our framework performed better than others by large margins on all metrics. The performance of our method demonstrates the better generalization ability of TriVol compared with the SOTA methods and baselines.

\subsection{Evaluate Object-Level Rendering}
\begin{figure*}[t]
	\centering
	\includegraphics[width=0.8\linewidth]{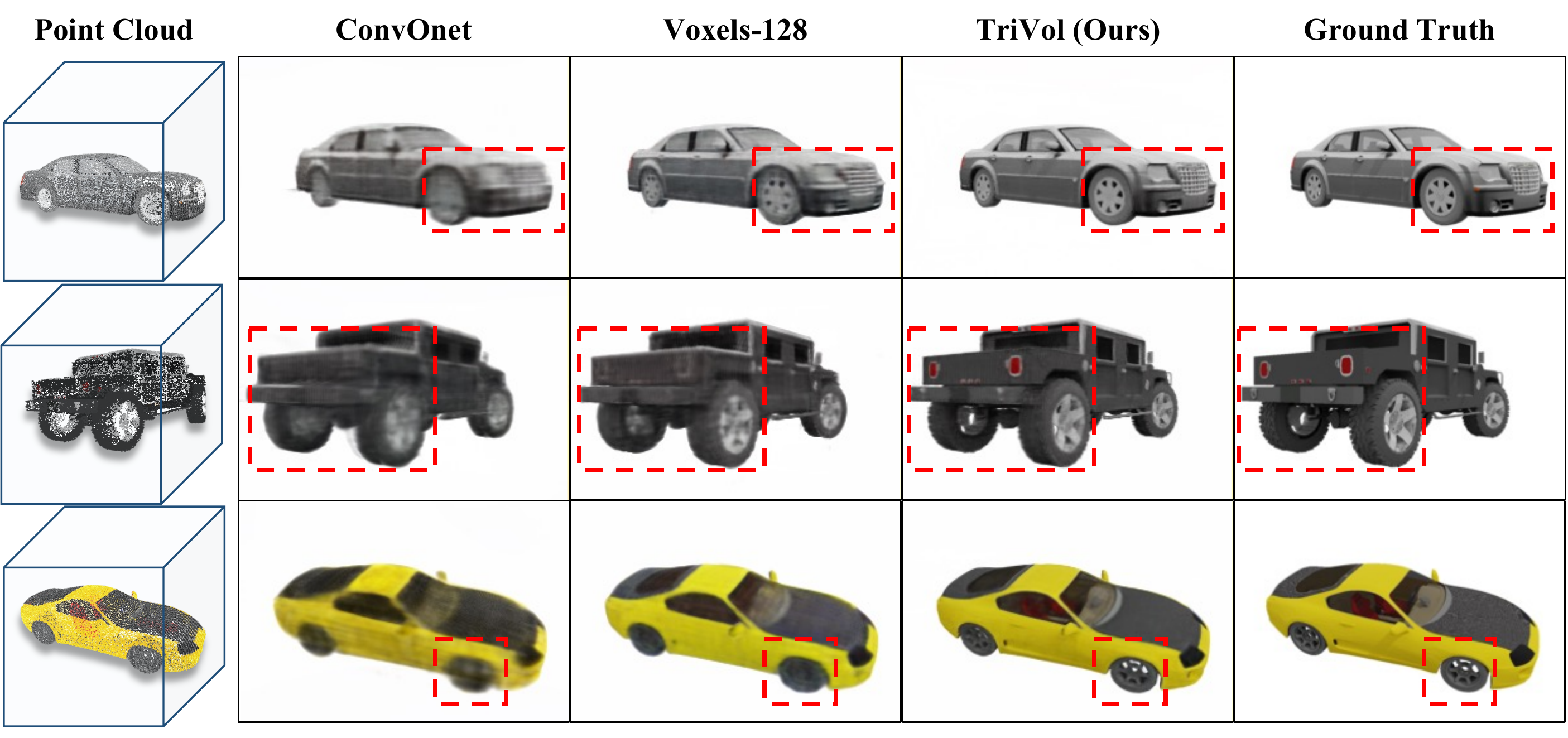}
	\caption{Qualitative point cloud rendering comparison between ours and SOTA methods and baselines on the ShapeNet dataset.}
    \label{fig:visualization-shapenet}
    \vspace{-0.2in}
\end{figure*}

\begin{figure*}[t]
	\centering
	\includegraphics[width=0.8\linewidth]{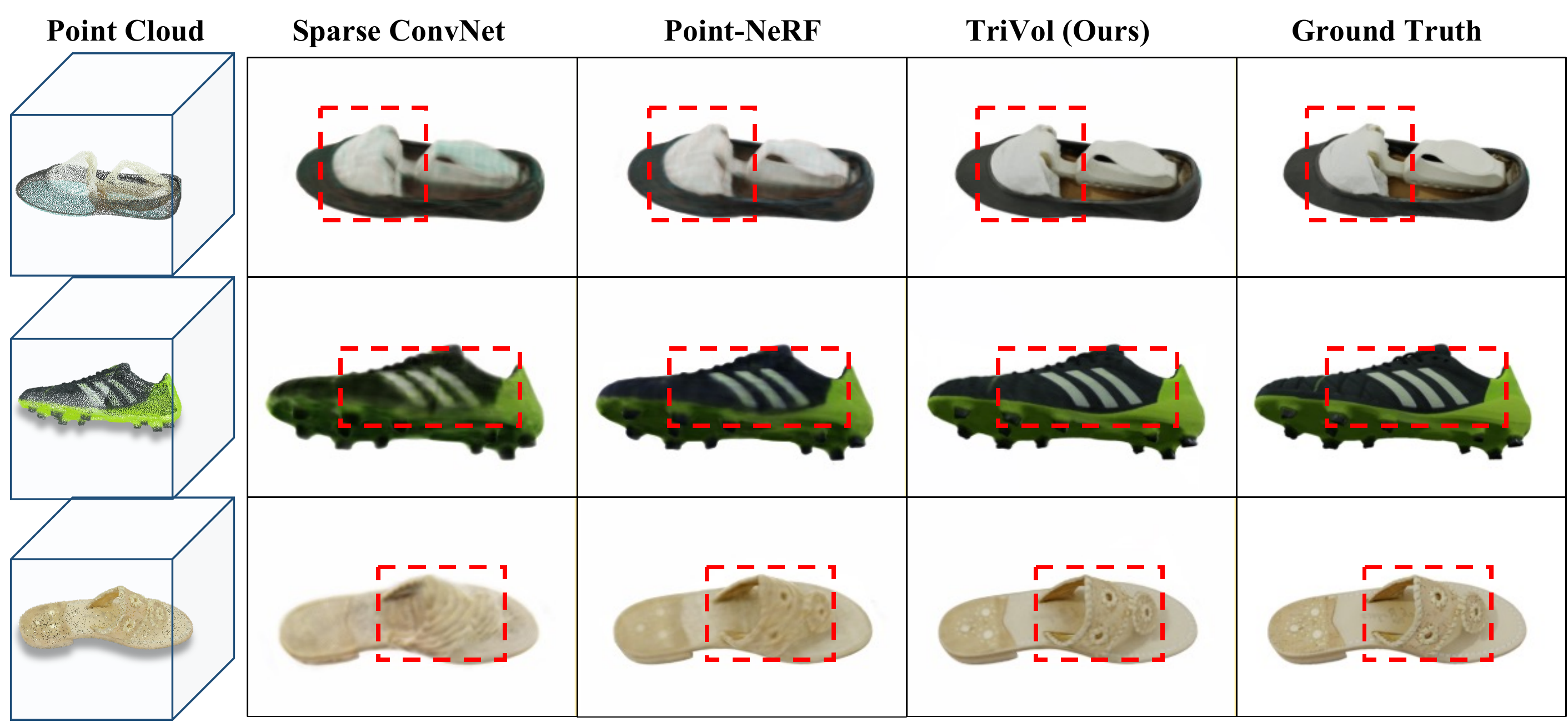}
	\caption{Qualitative point cloud rendering comparison between ours and SOTA methods and baselines on the GSO dataset.}
    \label{fig:visualization-gso}
\end{figure*}

We also conduct the evaluation of rendering at the object level.
Experiments are conducted on the ShapeNet \textit{Car} dataset and the Google Scanned Object \textit{Shoe} dataset. We uniformly sample 100k points for each 3D model. Different from the scene level, most of the point cloud at the object level is relatively dense. 
Therefore, all methods can achieve higher metrics than the results at the scene level.
The quantitative results are shown in Tab. \ref{tab:metric_comparison}, our framework still has the best performance over others, showing the discriminative 3D representation of TriVol.
Moreover, the qualitative evaluations are displayed in Figs. \ref{fig:visualization-shapenet} and \ref{fig:visualization-gso}.
Although there is no fine-tuning process, our method produces rendering results with richer and more precise details than baselines. For example, as shown in Fig.~\ref{fig:visualization-shapenet}, the car wheel structures could be accurately rendered with our approach, and are blurry at baselines' results.
The realistic rendering performance demonstrates the effectiveness of our framework in general category-specific point cloud rendering. Please refer to the supplementary file for more object-level comparisons within various categories.

\subsection{Ablation Study}
In the subsequent experiments, we study the influence of different modules and parameters in our method. 
\subsubsection{The Effect of Encoder}
In this experiment, we explore the performance of different encoders in transforming a point cloud to our \textit{Initial TriVol}. The baselines are the encoders containing the point-based networks, including PointNet \cite{Point-NeRF}, PointNet++  \cite{PointNet++}, and Sparse ConvNet \cite{ME}. 
After the point feature extraction in baselines, we pool the point features to our TriVol's shape.
The results are shown in Tab. \ref{tab:encoder}.  We note that our grouping strategy does not require additional memory while performing the best, showing the strength in employing the grouping mechanism in the TriVol encoder.


\begin{table}[t]
\begin{adjustbox}{width=1.0\columnwidth,center}
\begin{tabular}{c|cccc}
\hline
Encoder & Memory (GB) & PSNR ($\uparrow$) & SSIM ($\uparrow$) & LPIPS ($\downarrow$) \\ \hline
PointNet \cite{PointNet}  & 4.78 & 18.03 & 0.698 & 0.521 \\
PointNet++ \cite{PointNet++} & 5.23 & 18.26 & 0.707 & 0.505 \\
Sparse ConvNet \cite{ME} & 2.15 & 18.50 & 0.719 & 0.494 \\ \hline
\textbf{Grouping (Ours)} & \textbf{0} & \textbf{18.56} & \textbf{0.734} & \textbf{0.473}\\ \hline
\end{tabular}
\end{adjustbox}
\vspace{-0.1in}
\caption{The effect of different encoders on the ScanNet Dataset. The decoder of these methods are all the same three 3D UNet.}
\label{tab:encoder}
\vspace{-0.1in}
\end{table}

\subsubsection{The Effect of TriVol Resolution}
One advantage of our method is that high-resolution TriVol can be obtained with lightweight computation through dense 3D ConvNets.
In this experiment, we study the influence of TriVol resolution, including the group size $G$ and volume resolution $S$. 
The experimental results are shown in Tab. \ref{tab:trivol_resolution}, where ``Number" indicates the number of volumes in each 3D representation. 
Thus, the top two rows are the baselines of dense voxels with the resolution of $128^3$ and $256^3$, respectively.
The dense voxel with resolution $128^3$ requires a tremendous amount of computation resources in terms of GPU Memory and FLOPS. The voxels of $256^3$ even return an Out-Of-Memory (OOM) issue. 
Moreover, we noted that the rendering accuracy with these dense voxels is not satisfying enough.
On the other hand, the bottom four rows in Tab. \ref{tab:trivol_resolution} explore the effect of our framework with different combinations between group sizes and volume resolutions. Note that different combinations' memory costs are lower than the dense voxel with $128^3$ resolution. The superiority in the efficiency of TriVol can also be observed from the metric of FLOPS. Moreover, TriVol has an much better accuracy in rendering metrics.
Therefore, we finally adopt $G=16$, $S=256$ to balance the computing resources and accuracy, which has been indicated in Sec.~\ref{sec:details}.

\begin{table}[t]
\begin{adjustbox}{width=1.0\columnwidth,center}
\begin{tabular}{ccc|ccc}
\hline
Number & G & S & \makecell{GPU Memory \\ ($GB, \downarrow$)}  & FLOPS ($G, \downarrow$) & PSNR ($\uparrow$) \\ \hline
1 & - & 128$^3$ & 13.83 & 167.91 & 25.53 \\
1 & - & 256$^3$ & OOM & - & - \\ \hline
3 & 8 & 64$^2$ & 2.11 & 9.34 & 22.35 \\
3 & 16 & 128$^2$ & 3.13 & 74.69 & 24.84 \\
3 & 16 & 256$^2$ & 8.47 & 103.48 & \textbf{27.22} \\
3 & 32 & 256$^2$ & 9.89 & 197.55 & \textbf{27.38} \\ \hline
\end{tabular}
\end{adjustbox}
\caption{The effect of TriVol Resolution on the ShapeNet Dataset.} 
\label{tab:trivol_resolution}
\end{table}

\subsubsection{The Effect of Point Number}
In the real-world application, the testing scenes or objects might contain a visibly different point number from the training data, leading to performance degradation.
Therefore, in this experiment, we set different point numbers for the training and evaluation data, as shown in Tab. \ref{tab:number_of_points}.
When the point number during the training and testing are the same, similar to Point-NeRF \cite{Point-NeRF}, increasing the input point number can improve the rendering accuracy. Our method achieves the best performance with different numbers of input points. 
When the number of testing points is far less than the train points, our approach shows a more robust performance compared with Point-NeRF \cite{Point-NeRF} since TriVol is a general discriminative and continuous 3D representation regardless of the point cloud's sparse/dense degree.
Note that although the point number is changed, TriVol's resolution is fixed in this experiment.
Hence, our TriVol is more robust than Point-NeRF \cite{Point-NeRF} when the points' sparsity changes.

\begin{table}[t]
\begin{adjustbox}{width=1.0\columnwidth,center}
\begin{tabular}{c|ccc}
\hline
Methods & Training Points & Testing Points & PSNR ($\uparrow$) \\ \hline
\multirow{3}{*}{Point-NeRF \cite{Point-NeRF}} & 10k & 10k & 23.83 \\
 & 100k & 10k & 22.75 \\
 & 100k & 100k & 25.73 \\ \hline
\multirow{3}{*}{TriVol (Ours)} & 10k & 10k & \textbf{25.31} \\
 & 100k & 10k & \textbf{27.01} \\
 & 100k & 100k & \textbf{27.22} \\ \hline
\end{tabular}
\end{adjustbox}
\caption{The effect of the number of points during training and testing on the ShapeNet dataset.}
\label{tab:number_of_points}
\end{table}

\section{Conclusion}
\label{sec:conclusion}
In this paper, we analyze the limitations of present point-based rendering methods, including the hole artifacts for graphics-based methods, the view inconsistency for the 2D neural-based approaches, and the inefficiency of existing 3D neural-based strategies.
We propose a novel rendering framework by designing a lightweight and continuous 3D representation TriVol.
The \textit{Initial TriVol} is obtained from the point cloud with a simple axis grouping mechanism, and the \textit{Feature TriVol} is inferred by efficient computation in 3D UNets.
The feature querying for all 3D locations can be completed via the trilinear interpolation in TriVol.
The TriVol-based framework can represent high-resolution volumes and adaptive capture both local and non-local information for NeRF-based volume rendering.
Extensive experiments are conducted on both scene- and object-level benchmarks, showing our framework can be utilized in rendering images with clear details and without holes artifacts.

\noindent\textbf{Limitation and future work}. 
Although high-quality rendering images were obtained with TriVol from point clouds, it is still very challenging for our method to render the scene where an extremely large number of points are missing. This is technically a 3D scene synthesis task and requires a pre-trained 3D generator trained on a large-scale dataset to synthesize more missing points. 
We plan to solve it in future work based on the proposed TriVol representation. 

\section*{Acknowledgments}
This work is partially supported by Shenzhen Science and Technology Program KQTD20210811090149095.

{\small
\bibliographystyle{ieee_fullname}
\bibliography{cvpr}
}

\end{document}